\documentclass{article}
\usepackage{ijcai22}

\usepackage{times}

\usepackage{soul}
\usepackage{url}
\usepackage[hidelinks]{hyperref}
\usepackage[utf8]{inputenc}
\usepackage[small]{caption}
\usepackage{graphicx}
\usepackage{amsmath}
\usepackage{mathtools}
\usepackage{xcolor}
\usepackage{amssymb}
\usepackage{booktabs}
\usepackage{dsfont}
\usepackage{caption}
\usepackage{subcaption}
\usepackage{dblfloatfix}
\title{Text Transformations in Contrastive Self-Supervised Learning: A Review}

\author{
Amrita Bhattacharjee\footnote{Authors contributed equally to this work. Names are in alphabetical order.}\and
Mansooreh Karami$^*$\And
Huan Liu\\
\affiliations
Arizona State University, Tempe, AZ\\
\emails
\{abhatt43, mkarami, huanliu\}@asu.edu
}

\begin{document}

\maketitle

\begin{abstract}
Contrastive self-supervised learning has become a prominent technique in representation learning. The main step in these methods is to contrast semantically similar and dissimilar pairs of samples. However, in the domain of Natural Language Processing~(NLP), the augmentation methods used in creating similar pairs with regard to contrastive learning (CL) assumptions are challenging. This is because, even simply modifying a word in the input might change the semantic meaning of the sentence, and hence, would violate the distributional hypothesis. In this review paper, we formalize the contrastive learning framework, emphasize the considerations that need to be addressed in the data transformation step, and review the state-of-the-art methods and evaluations for contrastive representation learning in NLP. Finally, we describe some challenges and potential directions for learning better text representations using contrastive methods.

\end{abstract}

\section{Introduction}


Self-supervised learning uses the data itself to provide the supervisory signals for representation learning without any other costly annotating processes. This is valuable in many real-world scenarios nowadays where vast quantities of information are easily available but the cost of annotating such data is high. Based on the objective function of the deep neural networks, the self-supervised models can be divided into three major groups: generative, contrastive, and generative-contrastive~(or adversarial)~\cite{liu2021self}. 
In this paper, we focus on contrastive self-supervised models in NLP.
Unlike the generative models that apply the loss function on the output space, in contrastive models, the loss is measured in the representation space.

\begin{figure}[ht]
  \centering
  \includegraphics[width=1\linewidth]{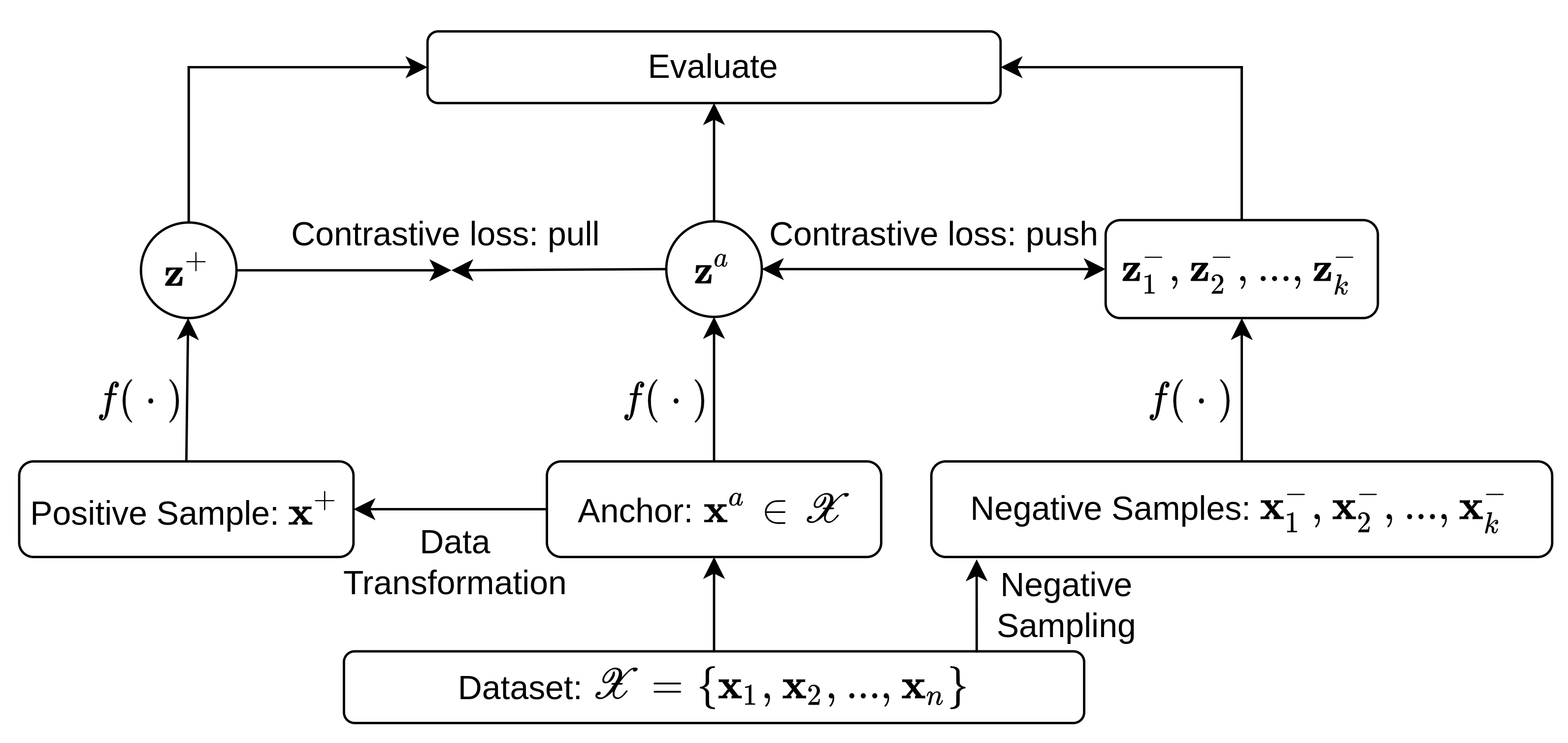}
  \caption{A learning step in CL and its effect on the representation space. The CL goal is to find a representation function (i.e., an encoder, $\mathbf{z}=f(\cdot)$) such that similar samples (i.e., the anchor, $\mathbf{x}^a$ , and the positive sample, $\mathbf{x}^+$)  are closer to each other and are pushed away from contrasting samples (negative-labeled instances).} 
  \label{fig:cl} 
\end{figure}

\begin{figure*}[ht]
    \centering
    \includegraphics[width=0.8\linewidth]{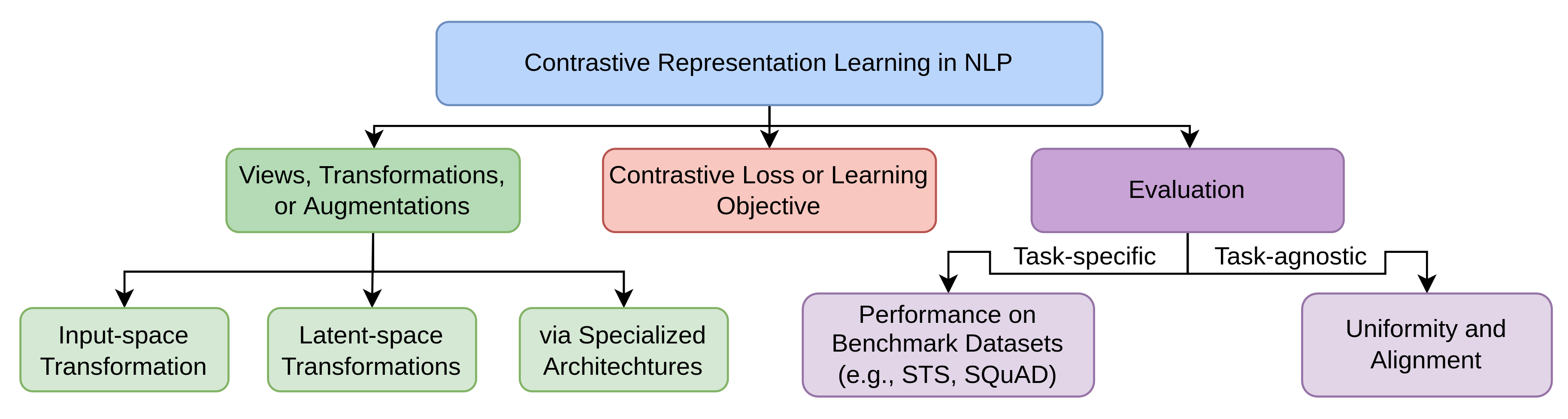}
    \caption{Taxonomy for Contrastive Learning on Text}
    \label{fig:taxonomy}
\end{figure*}
By creating pseudo-labels as supervision, the contrastive learning objective aims to bring the semantically similar samples close to each other and away from dissimilar instances. 
In a learning phase of a commonly used setup of CL in natural language, one sample from the training data acts as an anchor, its \textit{augmented} version is labeled as a positive sample, and the rest of the examples in the training batch are tagged as negative samples. An illustration of this learning step in CL is presented in Figure~\ref{fig:cl}. Unlike in images, the augmentation or transformation functions\footnote{In this paper, we use data `transformation' and data `augmentation' interchangeably.} used in creating the semantically similar pairs for texts are not well-defined and thus are more challenging. For example, in the task of word shuffling, `\textit{He had his car cleaned}' versus `\textit{He had cleaned his car}' has two different semantic implications and should not be used as similar pairs. On the other hand, a blind shuffle such as `\textit{cleaned He car his had}' does not conform to the grammar rules for English and should not be considered as a positive sample. To this means, we formalize the CL setup for NLP tasks~($\mathcal{x}$\ref{sec::setup}). We collect studies to present a representative survey of this field as shown in Figure \ref{fig:taxonomy}, specifically focusing on the different kinds of data augmentation used in creating the positive samples~($\mathcal{x}\ref{sec::aug}$) as well as sampling negative examples~($\mathcal{x}\ref{sec::neg_samples}$). We also review the different losses and evaluation metrics used in this area~($\mathcal{x}\ref{sec::loss}\text{ and } \mathcal{x}\ref{sec::eval}$). We conclude with open problems and challenges of the self-supervised CL for text representations and emphasize the considerations needed for choosing \textit{good} data transformations~($\mathcal{x}\ref{sec::chall})$.



Note that in this paper, we utilize bold lowercase letters for vectors (e.g., $\mathbf{x}$) as well as lowercase letters with and without input argument to represent functions (e.g., $f(\cdot)$) and scalars (e.g., $i$), respectively. Calligraphic letters are used for denoting sets and losses (e.g., $\mathcal{X}$) and Greek letters for parameters (e.g., $\beta$). To represent a sample from a set, we use subscript letters or numbers (e.g., $\mathbf{x}_k$ or $\mathbf{x}_1$). Uppercase letters in regular font is used for random variables (e.g., $U$). Finally, superscript $T$ denotes the transpose of a matrix or a vector.
\section{Contrastive Learning}
\label{sec::setup}

In this section, we will describe the contrastive learning (CL) framework. In CL, the aim is to learn a representation function $f:\mathcal{X}\rightarrow \mathcal{S}^{d-1}$ that maps all the data points $\mathcal{X}=\{\mathbf{x}_1, \mathbf{x}_2, ..., \mathbf{x}_n\}$ that are sampled from a distribution $p(\mathbf{x})$ to a hypersphere space in $\mathbb{R}^d$. Using commonly-used noise contrastive loss (NCE)~\cite{gutmann2010noise}, which is a lower bound on the mutual information of two random variables~\cite{poole2019variational}, the CL framework tries to pull the similar examples towards each other and push them away from the dissimilar examples.
This representation learning is conducted with an assumption that we have access to the similarity, $p^+$, and dissimilarity, $p^-$, distributions.
The quality of the learned representations in CL is highly dependent on informativity of the positive pairs $(\mathbf{x}^a, \mathbf{x}^+)$ and the negative samples $\{\mathbf{x}_1^-, \mathbf{x}_2^-, ..., \mathbf{x}_k^-\}$, where $\mathbf{x}^a \in \mathcal{X}$ is called anchor~\cite{robinson2020contrastive}. The samples should be paired such that they are semantically similar and dissimilar to $\mathbf{x}^a$ for the positive and negative data points.  

Following the formulation in~\cite{arora2019theoretical}, we formally define the concept of \textit{semantic similarity} by assuming a set of latent classes $\mathcal{C}$. Lets assume that $p(c)$ is a distribution over the latent classes that shows the natural occurrence of these classes in the unlabeled setting, where $c \in \mathcal{C}$. We also assume similar data points (i.e., $\mathbf{x}^a$ and $\mathbf{x}^+$) are i.i.d. and drawn from the same class distribution. Then, for some class $c$ sampled randomly from $p(c)$, the similarity and dissimilarity distributions are defined as:
\begin{equation}
\label{eq::dist_sim}
\small
    p^+(\mathbf{x}^a, \mathbf{x}^+) = \underset{c\sim p(c)}{\mathbb{E}}p(\mathbf{x}^a|c)p(\mathbf{x}^+|c)p(c)
\end{equation}
\begin{equation}
\label{eq::dist_dissim}
\small
    p^-(\mathbf{x}^-)=\underset{c\sim p(c)}{\mathbb{E}}p(\mathbf{x}^-| c)p(c),
\end{equation}
in which $\mathbf{x}^-$ is sampled i.i.d. from the marginal distribution of $p^+$.
Finally, the learning process will be maintained using the following loss:
\begin{equation}
\label{eq::nce_loss}
\small
\begin{multlined}
    \mathcal{I}(U; V) \geq \mathcal{L}_{\text{NCE}}(\mathbf{u}_i, \mathbf{v}_i)=\\\mathbb{E}_{(\mathbf{u}_i, \mathbf{v}_i) \sim p^+} \mathbb{E}_{\mathbf{v}_{1:k} \sim p^-} \left[\log \frac{e^{g(\mathbf{u}_i, \mathbf{v}_i)}}{\frac{1}{k+1}
    \underset{{j \in \{i, 1:k\}}}{\sum} e^{g(\mathbf{u}_i, \mathbf{v}_j)}} \right],
\end{multlined}
\end{equation}
where $\mathcal{I}(U; V)$ is the mutual information between two random variables $U$ and $V$, with $\mathbf{u}$ and $\mathbf{v}$ as their realizations, respectively. In our example in Figure~\ref{fig:cl}, $U$ and $V$ are derived from the same random variable sampled from the distribution $p(\mathbf{x})$. With $\mathbf{z} = f(\cdot)$ as our encoder, $g(\mathbf{u},\mathbf{v})$ can be defined as any similarity function between $f(\mathbf{u})$ and $f(\mathbf{v})$.


Current self-supervised CL approaches empirically try to follow the above setting, but face some challenges. First, they do not have access to the actual classes  and the similarity/dissimilarity information, so equations~(\ref{eq::dist_sim}) and~(\ref{eq::dist_dissim}) cannot be calculated directly. To this means, some heuristics have been applied to account for similarity such as user-specified transformation functions, data augmentation methods, and unsupervised clustering. On the other hand, negative instances are often sampled uniformly from the data, regardless of whether they share any semantic conformity with $x^a$ or not. In other words, if the selected instance as the negative is semantically similar to the anchor, their representation are still pushed apart. 
This sampling bias would lead to a sub-optimal representation as it cannot capture the true semantic structure of the data~\cite{li2020prototypical,chuang2020debiased}. 
Second, in the process of data augmentation, the i.i.d. assumption might not hold anymore. For example, in NLP, perturbing the instances in creating the positive samples might alter the semantics, change the distribution of the sample, and create a more negative pair rather than a positive one.
%
\begin{table*}[ht]
\centering
\fontsize{8.5}{9}\selectfont 
\begin{tabular}{|p{4.8cm}|p{4.8cm}|p{3.9cm}|p{2.5cm}|}
\hline
\textbf{Transformation or Augmentation} & \textbf{Loss} & \textbf{Tasks} & \textbf{Paper} \\
\hline\hline

BERT Siamese/Dual Encoder & Approximate nearest neighbor NCE & Dense text retrieval & \cite{xiong2020approximate} \\\hline
         Summarization of documents (no negative examples used) & NLL + multiple similarity losses & Abstractive text summarization & \cite{xu2021sequence} \\\hline
         Adversarial +ve/-ve examples by adding perturbations to the latent representation & combination of KL divergence and contrastive loss & Machine translation, text summarization, question generation & \cite{lee2020contrastive} \\\hline
         Spans of text sampled from the same document, i.e. anchor and positive from the same document & InfoNCE + MLM loss & SentEval benchmark tasks & \cite{giorgi2020declutr} \\\hline
         Word deletion, Span deletion, Reordering, Synonym substitution & n-pairs contrastive loss & GLUE  & \cite{wu2020clear} \\\hline
         Back-translation & n-pairs loss (based on momentum CL) & GLUE & \cite{fang2020cert} \\\hline
         Different dropout masks for anchor and positive & n-pairs loss & STS tasks and SICK-R~\cite{marelli2014sick} & \cite{gao2021simcse} \\\hline
         Adversarial perturbations, token shuffling, cutoff, dropout & n-pairs + cross-entropy for supervised task & STS tasks and SICK-R~\cite{marelli2014sick} & \cite{yan2021consert} \\\hline
         Inference relations from SNLI~\cite{bowman2015large} & supervised contrastive loss + cross-entropy & STS tasks and SentEval & \cite{liao2021sentence} \\\hline
         Corrupted and cropped sequences as positives & MLM + CLM + a form of n-pairs contrastive loss & GLUE and SQuAD & \cite{meng2021coco} \\\hline

\end{tabular}
\caption{Types of Transformations/Augmentations along with the loss function and evaluation task, over a set of representative works (NLL: Negative Log-Likelihood, MLM: Masked Language Modeling Loss, CLM: Corrective Language Modeling Loss)}
\label{tab:tasks}
\end{table*}
In this paper, we overview different heuristics used in defining the positive and negative samples in text-based CL.
Furthermore, we explain several caveats and hard assumptions implicit in standard CL frameworks, and hence, guide readers to delve into open problems in this area.

\section{Data Augmentation}
\label{sec::aug}



In a standard CL framework, the first step is to generate positive and negative samples for a given anchor data point. However, making transformations on the anchor to generate such positive samples is a more complicated task in the text domain, given the discrete nature of the input space. In this section, we review some commonly used transformation or augmentation methods used in different settings. Although a direct mapping between the downstream task and the appropriate augmentation is not straight-forward, a comparison of transformations, tasks, and losses over some representative works is given in Table \ref{tab:tasks}.




\subsection{Input-Space Transformations}


The most straight-forward class of text transformations would be operations performed in the discrete input space, also known as instance-based transformation. Even though these transformation methods are not as intuitive as similar transformations in image (such as cropping, flipping or rotating), different approaches have been explored in literature with varying degrees of success. 
In their DeCLUTR method, Giorgi et al.~\cite{giorgi2020declutr} used a \textit{span sampling} approach and considered segments that are adjacent to, overlapped with, or subsumed the original text segment, as positive samples. The augmented samples may also be created by \textit{lexical and sentence transformation}. Wu et al.~\cite{wu2020clear} used approaches such as word deletion, span deletion, token reordering, and synonym substitution for sentence augmentation in their CL method, CLEAR. Other token-level augmentation methods that have been proposed as standard data augmentation~\cite{wei2019eda}, such as synonym replacement, random insertion, random swap, and random deletion, can also be used for generating the positive pairs. A recent work in open domain question answering~\cite{ram2021learning} used cross-passage recurring spans of text as the positives - one span act as the anchor (or `query' in dense retrieval terms), while another acts as the positive. 

\subsection{Latent-Space Transformations}

Techniques proposed for standard data augmentation in low-resource learning settings can also be used to generate positive samples for contrastive representation learning in text. Some of these methods that generally preserve the semantic meaning of the original text include \textit{back-translation} using another intermediate language~\cite{fang2020cert,xie2019unsupervised} and \textit{language models} to replace selected words from the text with nearest neighbor words~\cite{jiao2019tinybert} such as word2vec~\cite{mikolov2013efficient} or GloVe. Xu et al.~\cite{xu2021sequence} utilized a document-level CL to train a document-level \textit{summarization model}. For the CL scheme, the authors use the original document, its gold summary, and the generated summary as different views of the data. The choice of these views is motivated by the idea that an article and its summarization must be close to each other in the semantic space.
Meng et al.~\cite{meng2021coco} used CL as part of their model for language model pretraining. The positive samples consist of the cropped version that keeps a random 90\% contiguous span of the original sentence and the recovered sentence from the \textit{masked language model} by randomly masking some words in the original sentence.

\subsection{Transformations via Architecture and Combined Methods}

Positive pairs for text may also be generated using slightly different architectures or modifying some aspect of the architecture in a certain way. One such architecture based method for text augmentation in CL utilizes \textit{dropout noise}. Gao et al.~\cite{gao2021simcse} create the positive pairs by feeding the sample input to the encoder twice and getting two embeddings with different dropout masks. The embeddings then is used as $i$ and $j$ samples in equation~(\ref{eq:contrastive_loss}). A perturbed version of the input can be generated by \textit{adversarial training} and tagged as a positive example. Yan et al.~\cite{yan2021consert} not only used lexical transformation and dropout approaches for data augmentation but also perturbed the input by applying Fast Gradient Value (FSV)~\cite{rozsa2016adversarial} as the adversarial attack method. Finally, some approaches consider \textit{inference relations} in Natural Language Inference datasets to create the desired data. These datasets consist of a premise-hypothesis pair with three different relationships: entailment, neutral, or contradiction. The premise acts as an anchor while the hypothesis would be labeled as positive if the relationship is entailment and negative if it is either neutral or contradiction~\cite{liao2021sentence}. 

\section{Negative Sampling}
\label{sec::neg_samples}

In previous sections, we review heuristics used to create the positive samples. Unlike studies in deep metric learning~\cite{suh2019stochastic}, the value of the negative samples has been understated in unsupervised contrastive representation learning.
Different samples of negative datapoints have different effects on the quality of the final representation. An efficient sampling function for these negative examples can also facilitate the learning process by correcting the model's mistake more quickly. Specifically, samples that are mapped near the anchor with high propensity in having the same label can significantly help in improving the representations. These samples are known as hard negatives. When latent classes are known (i.e., the supervised case), it is easy to identify task-specific hard negatives. But, in unsupervised settings, mining the hard negatives is more challenging. In these settings, researchers often increase the batch size such that the loss function covers a diverse set of negative samples~\cite{chen2020simple,he2020momentum}. However, beside the heavy burden of large memory usage, Arora et al.~\cite{arora2019theoretical} prove that due to the inherent nature of CL, large number of negative samples in some cases might even decrease the performance of the downstream task. To this means, researchers proposed various methods in sampling the negative examples.

Robinson et al.~\cite{robinson2020contrastive} proposed a simple method for finding hard negative samples. The authors contemplate two ways for sampling the negatives: (1)~they used heuristics to make sure that the anchor and the negative sample correspond to different latent classes, and (2)~the selection of the negative samples is regulated by the parameter $\beta$ that controls the degree of similarity to the anchor, $e^{\beta f(\mathbf{x}^a)^T f(\mathbf{x}^-)}$. In other words, $\beta$ would up-weight the negative points that have larger inner product (i.e., small Euclidean distance) with the anchor. They examined the effectiveness of their approach on learning meaningful representations for different tasks on images, graphs, and texts. Similarly, Wu et al.~\cite{wu2020conditional} showed that difficult samples drawn from their proposed restricted class of distributions would pick the ones that are more similar to the anchor, hence yielding a stronger representations. Tested only on visual transfer tasks, they also defined a conditional noise CL estimator that has a lower variance than the commonly-used CL losses.

Xiong et al.~\cite{xiong2020approximate} raised the issue of in-batch negative and hard negative sampling, as local negative sampling will lead to diminishing gradient norms, large stochastic gradient variances, and slow convergence. To overcome this problem, they proposed a new CL method named as Approximate nearest neighbor Negative Contrastive Estimation (ANCE) which selects the negative samples from the entire dataset using an asynchronously updated ANN index.
In the context of vision, Kalantidis~\cite{kalantidis2020hard} also raised the same issue with the in-batch negative sampling as well as the time-consuming use of memory banks that needs to keep a large memory up-to-date. The authors proposed the Mixing of Contrastive Hard negatives (MoCHi) approach to synthesize hard negative features, by creating convex linear combinations of the hardest existing negatives. Their experiments show that MoCHi is able to learn generalizable representations faster than the SOTA self-supervised approaches.
Comparably, Chuang et al.~\cite{chuang2020debiased} pointed out that the selected negative samples in traditional CL might suffer from the sampling bias which can lead to significant performance drop. They proposed an unsupervised debiased contrastive loss that corrects for the sampling of datapoints with the same label.
Giorgi et al.~\cite{giorgi2020declutr} used both easy and hard negative samples from text documents. Their definition of hard negative samples is those that are in the same document as the anchor while their text is not subsumed, overlapped, or adjacent to the anchor. However, this would not guarantee that they are not semantically unrelated.

\section{Contrastive Losses}
\label{sec::loss}

Although equation~(\ref{eq::nce_loss}) is a general form of the contrastive loss, several variations of contrastive loss function have been used so far. One of the earliest contrastive loss functions used was in the context of energy based model to measure similarities between faces for face verification~\cite{chopra2005learning}. For two data instances, this intuitive learning objective was intended to give a small value of the loss if the data instances were from the same class, and would give a large loss value if they are from different classes. Keeping convergence and training efficiency in mind, over time, different variants of contrastive loss functions have been proposed.

In a self-supervised manner, the contrastive loss for a pair of positive samples $\mathbf{x}^a$ and $\mathbf{x}^+$ is calculated as follows:
\begin{equation}
\small
\begin{aligned}
\mathcal{L}(\mathbf{x}^a, \mathbf{x}^+)=-\log{\frac{e^{\text{sim}(f(\mathbf{x}^a), f(\mathbf{x}^+))/\tau}}{\sum_{k=1}^{2n}\mathds{1}_{[\mathbf{x}^a\neq \mathbf{x}_k^-]}e^{\text{sim}(f(\mathbf{x}^a), f(\mathbf{x}_k^-))/\tau}}},
\label{eq:contrastive_loss}
\end{aligned}
\end{equation}
where $n$ is the number of samples in one batch, $\mathds{1}_{[\mathbf{x}^a\neq \mathbf{x}_k^-]} \in \{0, 1\}$ is an indicator function, $\tau$ denotes a temperature hyperparameter, and $\text{sim}(\mathbf{u},\mathbf{v}) = \frac{\mathbf{u}^T\mathbf{v}}{||\mathbf{u}||\cdot||\mathbf{v}||}$ is the cosine similarity between two vectors.

Triplet loss \cite{schroff2015facenet} uses a triplet of an anchor, a positive sample~(i.e. has the same label as the anchor) and a negative sample~(i.e. different label from the anchor). This loss (shown in equation~(\ref{eq:trip})) tries to minimize the distance between the anchor and the positive and increase the distance between the anchor and the negative. 
\begin{equation}
\small
    \begin{aligned}
    \mathcal{L}(\mathbf{x}^a, \mathbf{x}^-, \mathbf{x}^+) = \max(||f(\mathbf{x}^a)-f(\mathbf{x}^+)||^2\\
    - ||f(\mathbf{x}^a)-f(\mathbf{x}^-)||^2 + \alpha,0),
    \end{aligned}
    \label{eq:trip}
\end{equation}

\noindent
where $\mathbf{x}^a$ is the anchor datapoint, $\mathbf{x}^-$ is the negative sample, $\mathbf{x}^+$ is the positive sample, and $\alpha$ is the margin between positive and negative samples. 
Similar approaches for learning the distance metric between instances have been proposed in other works such as \cite{schultz2004learning,wang2014learning}. Apart from the direct task of distance metric learning, contrastive loss has also been used for dimensionality reduction \cite{hadsell2006dimensionality}. However, one major problem with the triplet loss and several other similar variants of contrastive losses is that of \textit{hard negative mining}. The model would successfully learn the distance between positive and negative samples if the triplets are selected and constructed properly. However, as we explained in the previous section, searching for hard negatives over the entire training dataset is infeasible in practice, hence there have been several interesting approaches to solve this (as listed in $\mathcal{x}$\ref{sec::neg_samples}). 
Oh Song et al.~\cite{oh2016deep} propose the lifted structure loss that results in more efficient training and stable optimization.
\begin{equation}
\label{eq:lift}
\small
\begin{multlined}
    \mathcal{L}(\mathbf{x}^a,\mathbf{x}^+) =\underset{(\mathbf{x}^a,\mathbf{x}^+) \sim p^+}{\mathbb{E}} \max(0,\mathcal{J}(\mathbf{x}^a,\mathbf{x}^+))^2,\\
    \mathcal{J}(\mathbf{x}^a,\mathbf{x}^+) = \max[\max_{\mathbf{x}^a,\forall {\mathbf{x}^-}} \alpha - d(\mathbf{x}^a,\mathbf{x}^-),\\ \max_{x^+,\forall {\mathbf{x}^-}} \alpha - d(\mathbf{x}^+,\mathbf{x}^-) ] + d(\mathbf{x}^a,\mathbf{x}^+)
\end{multlined}
\end{equation}

\noindent
where $d(\mathbf{u},\mathbf{v}) = ||f(\mathbf{u})-f(\mathbf{v})||_2$ is the L2 distance between the representations of $\mathbf{u}$ and $\mathbf{v}$, and $\alpha$ is a margin parameter. This utilizes all the positive and negative pairs in a training batch. Furthermore, this approach tries to improve the representation learned, by looking for `difficult' negatives for a set of randomly chosen positive samples. 
Another issue with triplet loss and contrastive loss is that, especially for multi-class cases, it results in unstable updates and slow convergence. This is because in each step, only one comparison is being made with only one negative sample. This slows down the convergence. To alleviate this problem, \cite{sohn2016improved} proposed the n-pairs loss (equation~(\ref{eq:n-pair})), where in each step the loss is computed using a $(n+1)$ length tuple. 
\begin{equation}
\small
    \begin{multlined}
    \mathcal{L}(\{\mathbf{x}^a, \mathbf{x}^+, \{\mathbf{x}_i^-\}_{i=1}^{n-1}\})= \\ \log\left(1+\sum_{i=1}^{n-1}\exp\left(f(\mathbf{x}^a)^T f(\mathbf{x}_i^-) - f(\mathbf{x}^a)^Tf(\mathbf{x}^+)\right)\right)
    \end{multlined}
    \label{eq:n-pair}
\end{equation}

\noindent
where $f(\cdot)$ is the embedding kernel of the deep neural network (i.e., the encoder). This has one anchor element $\mathbf{x}^a$, one positive element $\mathbf{x}^+$, and $(n-1)$ negative samples $\{\mathbf{x}_i\}_{i=1}^{n-1}$. The n-pairs loss may be thought of as a special case of the lifted structure loss, where the batch contains positive pairs from disjoint classes. Furthermore, unlike the n-pairs loss, the lifted structure loss uses a max-margin based distance function in the loss formulation.

\begin{figure}[ht!]
    \centering
    \begin{subfigure}[b]{0.45\textwidth}
         \centering
         \includegraphics[width=0.45\textwidth]{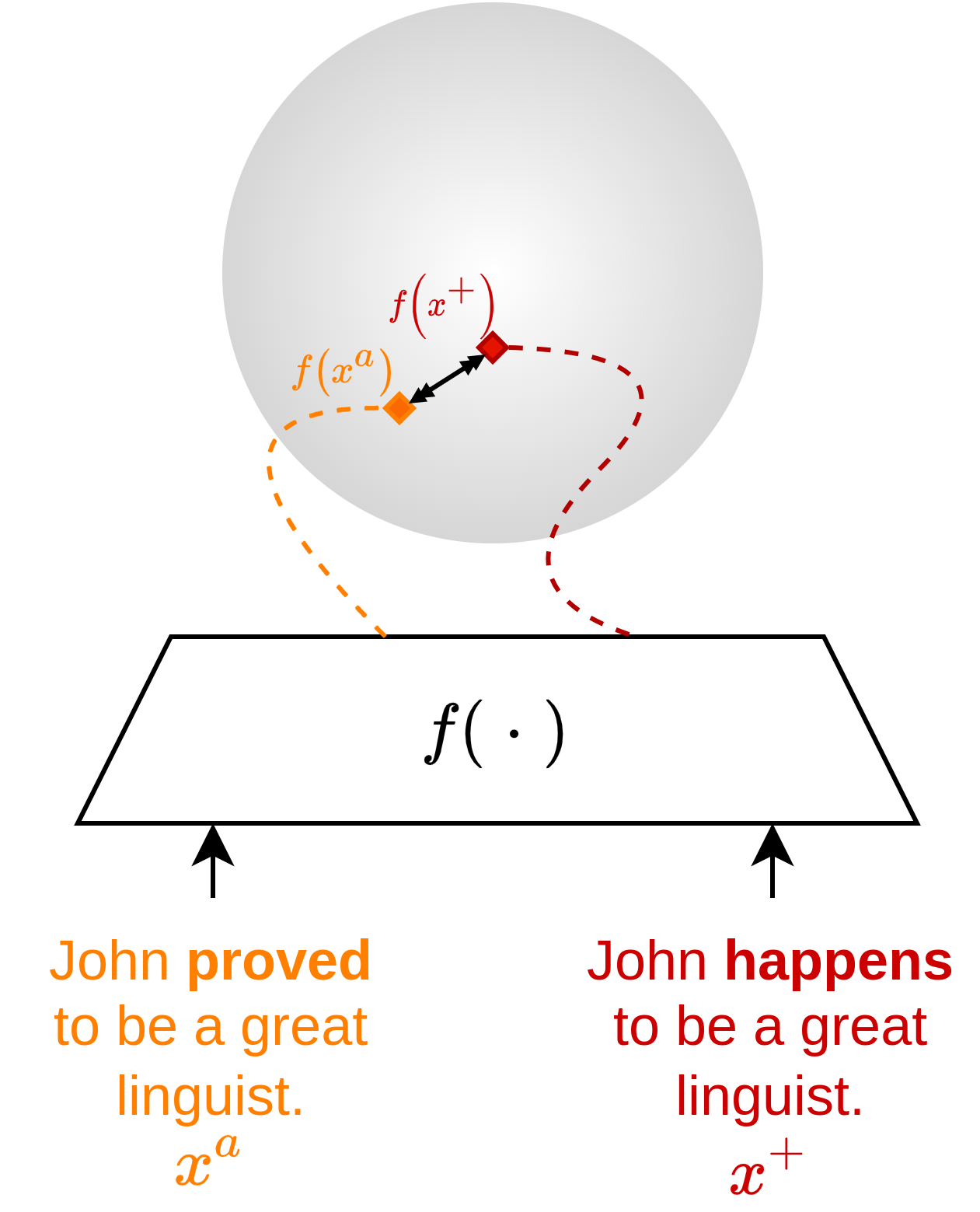}
         \caption{Closeness of features of positive pairs, $(\mathbf{x}^a,\mathbf{x}^+)\sim p^+$, on the hypersphere space.}
         \label{subfig:align}
     \end{subfigure}
     \hfill
     \begin{subfigure}[b]{0.45\textwidth}
         \centering
         \includegraphics[width=0.55\textwidth]{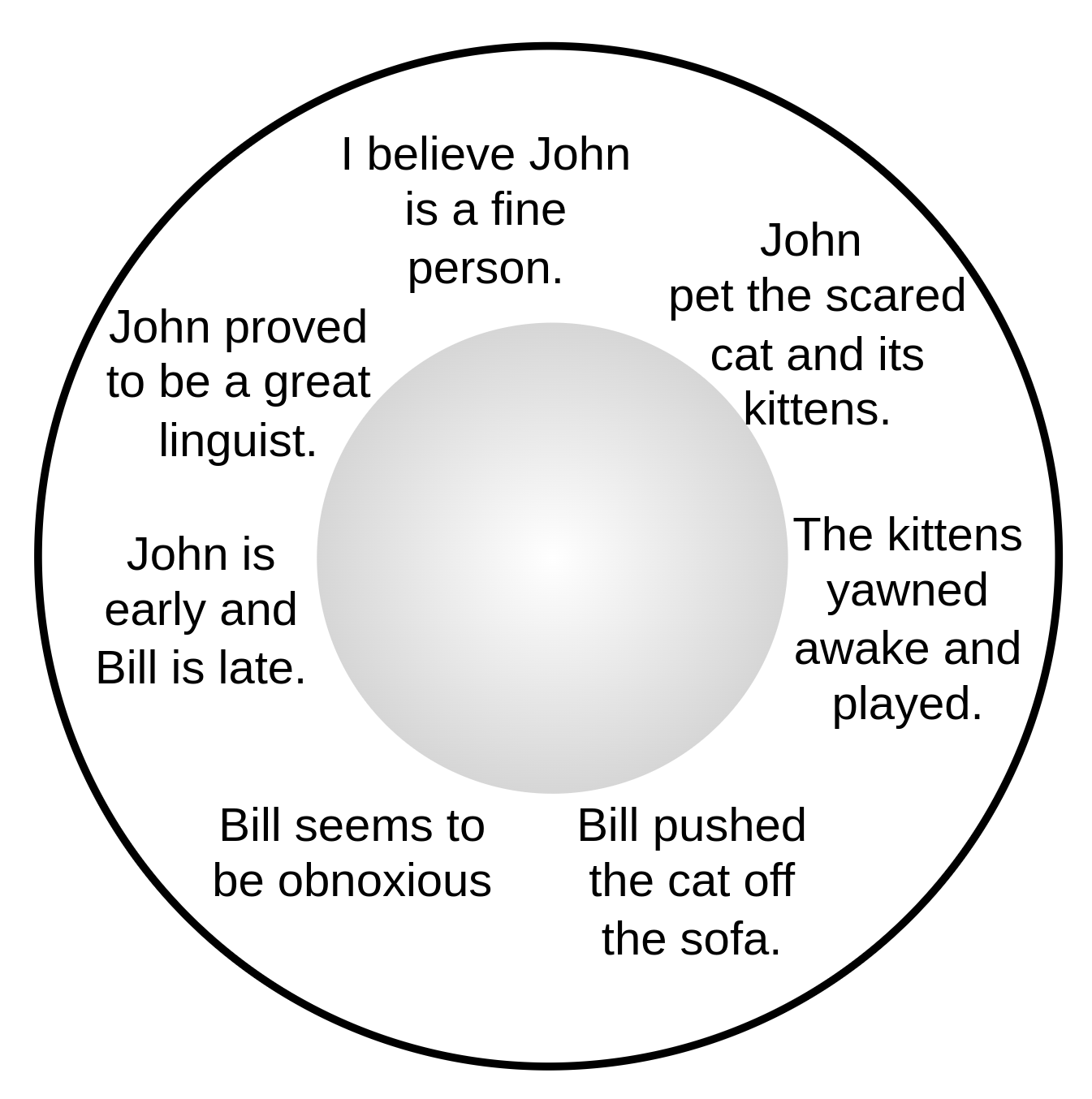}
         \caption{Uniformity in the distribution of features on a hypersphere.}
         \label{subfig:uniform}
     \end{subfigure}
    \caption{An illustration of (a)~Alignment and (b)~Uniformity of text representations on the output unit hypersphere. This figure is inspired by Wang and Isola's ~\protect\cite{wang2020understanding} illustration of these metrics on images. The sentences are extracted from the GLUE dataset~\protect\cite{wang2019glue} with some modifications.}
    \label{fig:eval}
\end{figure}

\section{Evaluation Metrics}
\label{sec::eval}

Most of the methods rely on the performance of the downstream tasks (e.g., accuracy on labeled benchmark datasets), in order to evaluate the quality of the learned representations. Moreover, for some specific objectives such as measuring the balancedness of a feature space, they rely on the linear separability performance which is evaluated by the accuracy of a linear classifier over the representation vectors~\cite{kang2020exploring,jiang2021improving}.
However, in learning \textit{universal} representations and in an \textit{unsupervised} manner, we can evaluate the quality of the representation by measuring how well the CL method separate similar pairs from dissimilar samples.   
Wang and Isola ~\cite{wang2020understanding} proposed two properties related to contrastive loss in assessing the CL representations. Since CL aims to find a representation space that the information is most shared between positive pairs as well as invariant to other noise factors, they defined these metrics:

\begin{itemize}
    \item[-] Alignment: the anchor and positive sample representations on the hypersphere space ($\mathcal{S}^{d-1}$) should be aligned and close to each other (Figure~\ref{subfig:align}), i.e., the absolute distance of the anchor and positive sample representation should be as small as possible: 
    \begin{equation}
    \small
    \label{eq::eval_align}
        \mathcal{L}_{\text{align}}(f; \rho)=\underset{(\mathbf{x}^a,\mathbf{x}^+)\sim p^+}{\mathbb{E}}\left[\parallel f(\mathbf{x}^a)-f(\mathbf{x}^+)\parallel_2^{\rho}\right], \rho>0
    \end{equation}
    \item[-] Uniformity: the distribution of the representations should roughly be uniform in the hypersphere space to preserve as much information of the data as possible (Figure~\ref{subfig:uniform}). This can be calculated as the logarithm of the average pairwise Gaussian potential kernel (also known as the Radial Basis
    Function (RBF) kernel), with parameter $\gamma$, between the representations of the data points $\mathcal{X}$:
    \begin{equation}
    \label{eq::eval_uniform}
    \small
        \mathcal{L}_{\text{uniform}}(f; \gamma)=\log
        \underset{(\mathbf{x}_i, \mathbf{x}_j)\stackrel{\scalebox{.5}{$   i.i.d$}}{\sim}\mathcal{X}}{\mathbb{E}}\left[ e^{-\gamma\parallel f(\mathbf{x}_i)-f(\mathbf{x}_j)\parallel_2^2}\right]
    \end{equation}
\end{itemize}
The authors also empirically showed that both $\mathcal{L}_{\text{align}}$ and $\mathcal{L}_{\text{uniform}}$ are strongly agree with and causally affect downstream task performance.





\section{Challenges and Open Problems}
\label{sec::chall}


\noindent
Alongside the success of CL in unsupervised settings, there has also been community-wide discussions regarding the generalizability of the representations learned by such methods, the appropiateness of the transformations, and several other related issues. In this section, we go deeper into some of the main challenges in contrastive self-supervised learning for text and point readers to potential directions for future research. 


\paragraph{The Selection of a Good Transformation Function.} Contrastive representation learning in the self-supervised setting assumes that the transformations that are done on the data points are semantically invariant, and hence are simply two `views' of the instance. Ideally the transformations or augmentations performed should not alter the semantic meaning of the data point. Most contrastive representation learning schemes assume that the downstream task that uses the learned representations would be invariant to the transformation performed during the learning process. For example, as explained in \cite{xiao2020should}, for a downstream task that does fine-grained classification of bird species, augmentations that involve modifying the color and texture of the image should not be performed, as these are useful features in identifying the species of bird. Similarly for text, augmentations that change the tone or sentiment of the sentence should not be used for learning representations in a system that does sentiment classification as the downstream task. A recent effort in this direction for images tries to learn invariant representations \cite{misra2020self}. In text, apart from the downstream task, the suitable transformation may also depend on the language.


\paragraph{Negative Samples and Sampling Bias.}
In the supervised counterpart of CL~\cite{khosla2020supervised}, sampling negative examples from truly different classes has shown to improve the performance of the representations. However, as mentioned in $\mathcal{x}$\ref{sec::neg_samples}, due to CL's unsupervised manner and the lack of access to the labels, we might accidentally sample false negatives and accept examples that are in reality semantically similar to the anchor. Future work is needed to mitigate this sampling bias without relying on the actual labels of the data.


\paragraph{Counterfactually-Augmented Data as Positive and Negative Samples.} Counterfactual examples have long been utilized and known to be useful for training, evaluating, and improving NLP models~\cite{moraffah2020causal,morris2020textattack} as well as mitigating bias~\cite{maudslay2019s,kaushik2019learning,hu2021causal}. By making sure that the counterfactual examples are plausible and not out-of-distribution to models~\cite{hase2021out}, there is a potential in creating augmented data that estimates the latent class for positive or negative examples and satisfies the assumptions for similarity and dissimilarity distributions. For example, by utilizing the relations between pairs of counterfactual examples we are able to find what changes in the input space are related to the change in the label~\cite{teney2020learning}. This technique, that is known as counterfactual data augmentation, seeks to eliminate spurious correlations using causal interventions~\cite{kaushik2019learning}. Moreover, label-preserving data augmentations can be used in generating examples that are similar to the anchor~\cite{joshi2021investigation}. 
Current generation methods mostly rely on human expert annotators to create the counterfactually-altered data in which they only instantiate limited types of perturbations like word substitutions. Methods such as Polyjuice~\cite{wu2021polyjuice} are attempts in automatically creating fluent and diverse counterfactual examples which support various downstream tasks on different domains. However, while using counterfactual examples in the context of CL, we have to make sure that the assumptions, such as preserved latent classes, identical distribution for positive samples, and the dissimilarity distribution requirements are not violated.

\paragraph{Euclidean vs non-Euclidean Spaces.} Most of the self-supervised NLP representation models such as word2vec~\cite{mikolov2013efficient}, GloVe~\cite{pennington2014glove}, and skip-thought vectors~\cite{kiros2015skip} are trained in the Euclidean space which aim to find a representation such that the distance between the vectors corresponds to their semantic proximity. Non-Euclidean spaces have also been explored for the purpose of the text representations. For example, Nickel and Kiela~\cite{nickel2017poincare} proposed a Poincar\'{e} embedding by utilizing the hyperbolic geometry for learning the similarity and the hierarchy of objects in predicting lexical entailment. Similarly, Dhingra et al.~\cite{dhingra2018embedding} showed that learning a Poincar\'{e} embedding for hierarchical structures will lead to an improvement on other downstream tasks and provided some evidence on the intuition of the hyperbolic embedding for structural data. 
Moreover, Meng et al.~\cite{meng2019spherical} showed that the spherical text embeddings would intrinsically capture the directional similarity. They proposed a model that would jointly learn word and paragraph embeddings. Prior to that, Batmanghelich et al.~\cite{batmanghelich2016nonparametric} applied von Mises-Fisher distribution to model the density of the words over a unit sphere as well as discovering the number of topics in the data. 
When all is said and done, the question of which space would capture the natural representation of the text is not yet rigorously answered. One possible direction is to critique the intuitions behind using the Euclidean, Hyperbolic, or Spherical spaces and provide evidence on the smoothness of the decision boundaries.

\section*{Acknowledgments}

This research is supported by the DARPA (HR001120C0123) and ONR (N00014-21-1-4002). The views, opinions and/or findings expressed are those of the authors and should not be interpreted as representing the official views or policies of the Department of Defense or the U.S. Government.

\bibliographystyle{named.bst}
\bibliography{reference.bib}

\end{document}